\documentclass[11pt,a4paper]{article}
\usepackage[hyperref]{acl2020}
\usepackage{times}
\usepackage{latexsym}

\usepackage{graphicx}
\usepackage{fancyvrb}
\usepackage{soul}
\usepackage{alltt}
\usepackage{subcaption}
\usepackage{multirow}

\usepackage{microtype}

\aclfinalcopy 

\definecolor{qhbrackets}{RGB}{228,26,28}
\definecolor{qhconstrainkey}{RGB}{152,78,163}
\definecolor{qhconstrainvalue}{RGB}{77,175,74}
\definecolor{qhname}{RGB}{55,126,184}
\definecolor{qhop}{RGB}{255,127,0}

\newcommand{\tgn}[1]{\textcolor{qhname}{#1}} 
\newcommand{\tgv}[1]{\textcolor{qhconstrainvalue}{#1}} 
\newcommand{\tgo}[1]{\textcolor{qhop}{#1}} 
\newcommand{\tgc}[1]{\textcolor{qhconstrainkey}{#1}} 

\newcommand{\query}[1]{{\small`\textsf{#1}'}}
\newcommand{\wquery}[1]{{\small\textsf{#1}}} 

\newcommand{\CAP}{\tgo{\textbf{:}}}
\newcommand{\expand}{\langle\rangle}

\title{Interactive Extractive Search over Biomedical Corpora}

\author{Hillel Taub-Tabib$^1$ \hspace{1em} Micah Shlain$^{1,2}$ \hspace{1em} \hspace{1em} \hspace {1em} Shoval Sadde$^1$ \hspace{1em} Dan Lahav$^3$ \\
\hspace {1em} \textbf{Matan Eyal}$^1$ \hspace{1em} \textbf{Yaara Cohen}$^1$ \hspace{1em} \textbf{Yoav Goldberg}$^{1,2}$ \\
$^1$ Allen Institute for AI, Tel Aviv, Israel \\
$^2$ Bar Ilan University, Ramat-Gan, Israel \\
$^3$ Tel Aviv University, Tel-Aviv, Israel \\
\texttt{\{hillelt,micahs\}@allenai.org}
\texttt{}
}

\date{}

\begin{document}
\maketitle
\begin{abstract}

We present a system that allows life-science researchers to search a linguistically annotated corpus of scientific texts using patterns over dependency graphs, as well as using patterns over token sequences and a powerful variant of boolean keyword queries. In contrast to previous attempts to dependency-based search, we introduce a light-weight query language that 
does not require the user to know the details of the underlying linguistic representations, and instead to query the corpus by providing an example sentence coupled with simple markup.
Search is performed at an interactive speed due to efficient linguistic graph-indexing and retrieval engine. This allows for rapid exploration, development and refinement of user queries.
We demonstrate the system using example workflows over two corpora: the PubMed corpus including 14,446,243 PubMed abstracts and the CORD-19 dataset\footnote{https://pages.semanticscholar.org/coronavirus-research}, a collection of over 45,000 research papers focused on COVID-19 research. The system is publicly available at \url{https://allenai.github.io/spike}
\end{abstract}

\section{Introduction}

Recent years have seen a surge in the amount of accessible Life Sciences data.
Search engines like Google Scholar, Microsoft Academic Search or Semantic Scholar allow researchers to search for published papers based on keywords or concepts, but search results often include thousands of papers and extracting the relevant information from the papers is a problem not addressed by the search engines.
This paradigm works well when the information need can be answered by reviewing a number of papers from the top of the search results. However, when the information need requires extraction of information nuggets from many papers (e.g. \emph{all chemical-protein interactions} or \emph{all risk factors for a disease}) the task becomes challenging and researchers will typically resort to curated knowledge bases or designated survey papers in case ones are available.

We present a search system that works in a paradigm which we call Extractive Search, and which allows rapid information seeking queries that are aimed at extracting facts, rather than documents.
Our system combines three query modes:
boolean, sequential and syntactic, targeting different stages of the analysis process, and different extraction scenarios.
Boolean queries (\S\ref{sec:boolean}) are the most standard, and look for the existence of search terms, or groups of search terms, in a sentence, regardless of their order. These are very powerful for finding relevant sentences, and for co-occurrence searches. Sequential queries (\S\ref{sec:sequential}) focus on the order and distance between terms. They are intuitive to specify and are very effective where the text includes ``anchor-words'' near the entity of interest. Lastly, syntactic queries (\S\ref{sec:query-by-example}) focus on the linguistic constructions that connect the query words to each other.  Syntactic queries are very powerful, and can work also where the concept to be extracted does not have clear linear anchors. However, they are also traditionally hard to specify and require strong linguistic background to use. Our systems lowers their barrier of entry with a specification-by-example interface.

Our proposed system is based on the following components.

\noindent\textbf{Minimal but powerful query languages.}
There is an inherent trade-off between simplicity and control. On the one extreme, web search engines like Google Search offer great simplicity, but very little control, over the exact information need. On the other extreme, information extraction pattern-specification languages like UIMA Ruta offer great precision and control, but also expose a low-level view of the text and come with over hundred-page manual.\footnote{\url{https://uima.apache.org/d/ruta-current/tools.ruta.book.pdf}} 

Our system is designed to offer \emph{high degree of expressivity}, while remaining simple to grasp: 
the syntax and functionality can be described in a few paragraphs. The three query languages are designed to share the same syntax to the extent possible, to facilitate knowledge transfer between them and to ease the learning curve.

\noindent\textbf{Linguistic Information, Captures, and Expansions.}
Each of the three query types are linguistically informed, and the user can condition not only on the word forms, but also on their lemmas, parts-of-speech tags, and identified entity types. The user can also request to \emph{capture} some of the search terms, and to \emph{expand} them to a linguistic context.
For example, in a boolean search query looking for a sentence that contains the lemmas ``treat" and ``treatment" (\query{\tgc{lemma}=\tgv{treat}\tgo{$|$}\tgv{treatment}}), a chemical name (\query{\tgc{entity}\tgo{=}\tgv{SIMPLE\_CHEMICAL}}) and the word ``infection" (\query{\tgv{infection}}), a user can mark the chemical name and the word ``infection" as \emph{captures}.
This will yield a list of chemical/infection pairs, together with the sentence from which they originated, all of which contain the words relating to treatments. Capturing the word ``infection'' is not very useful on its own: all matches result in the exact same word. But, by \emph{expanding} the captured word to its surrounding linguistic environment, the captures list will contain terms such as ``PEDV infection", ``acyclovir-resistant
HSV infection" and ``secondary bacterial infection".
Running this query over PubMed allows us to create a large and relatively focused list in just a few seconds. The list can then be downloaded as a CSV file for further processing.
The search becomes \emph{extractive}: we are not only looking for documents, but also, by use of captures, \emph{extract information} from them.

\noindent\textbf{Sentence Focus, Contextual Restrictions.}
As our system is intended for extraction of information, it works at the sentence level. However, each sentence is situated in a context, and we allow secondary queries to condition on that context, for example by looking for sentences that appear in paragraphs that contain certain words, or which appear in papers with certain words in their titles, in papers with specific MeSH terms, in papers whose abstracts include specific terms, etc. This combines the focus and information density of a sentence, which is the main target of the extraction, with the rich signals available in its surrounding context.

\noindent\textbf{Interactive Speed.} Central to the approach is an indexed solution, based on \cite{odinson}, that allows to perform all types of queries efficiently over very large corpora, while getting results almost immediately. This allows the users to interactively refine their queries and improve them based on the feedback from the results. This contrasts with machine learning based solutions that, even neglecting the development time, require substantially longer turnaround times between query and results from a large corpus. 

\section{Existing Information Discovery Approaches}
The primary paradigm for navigating large scientific collections such as MEDLINE/PubMed\footnote{\url{https://www.ncbi.nlm.nih.gov/pubmed/}} is document-level search.

The most immediate document-level searching technique is boolean search (``keyword search'').
However, these methods suffer from an inability to capture the concepts aimed for by the user, as biomedical terms may have different names in different sub-fields and as the user may not always know exactly what they are looking for. To overcome this issue several databases offer semantic searching by exploiting MeSH terms that indicate related concepts. While in some cases MeSH terms can be assigned automatically, e.g \cite{MTI-Mork}, in others obtaining related concepts require a manual assignment which is laborious to obtain.

Beyond the methods incorporated in the literature databases themselves, there are numerous external tools for biomedical document searching. Thalia \cite{Thalia} is a system for semantic searching over PubMed. 
It can recognize different types of concepts occurring in Biomedical abstracts, and additionally enables search based on abstract metadata; LIVIVO \cite{Livivo-2017} takes the task of vertically integrating information from divergent research areas in the life sciences; SWIFT-Review \footnote{\url{https://www.sciome.com/swift-review/}} offers iterative screening by re-ranking the results based on the user's inputs. 

All of these solutions are focused on the document level, which can be limiting: they often surface hundreds of papers or more, requiring careful reading, assessing and filtering by the user, in order to locate the relevant facts they are looking for.

To complement document searching, some systems facilitate automatic extraction of biomedical concepts, or patterns, from documents. Such systems are often equipped with analysis capabilities of the extracted information.
For example, NaCTem has created systems that extract biomedical entities, relations and events.\footnote{\url{http://www.nactem.ac.uk/}}; ExaCT and RobotReviewer \cite{ExaCT, RobotReviewer} take a RCT report and retrieve sentences that match certain study characteristics.

To improve the development of automatic document selection and information extraction the BioNLP community organized a series of shared tasks \cite{bionlp09,bionlp11,bionlp13, semeval-ddi, bionlp16bacteria, bionlp16seed, emnlp-2019-bionlp}. The tasks address a diverse set of biomed topics addressed by a range of NLP-based techniques.
While effective, such systems require annotated training data and substantial expertise to produce. As such, they are restricted to several ``head" information extraction needs, those that enjoy a wide community interest and support. The long tail of information needs of ``casual'' researchers remain mostly un-addressed.

\section{Interactive IE Approach}
Existing approaches to information extraction from bio-medical data suffer from significant practical limitations. Techniques based on supervised training require extensive data collection and annotation \cite{bionlp09,bionlp11,bionlp13, semeval-ddi, bionlp16bacteria, bionlp16seed}, or a high degree of technical savviness in producing high quality data sets from distant supervision \cite{nary-re,verga2017attending,suppai}. On the other hand, rule based engines are generally too complex to be used directly by domain experts and require a linguist or an NLP specialist to operate. Furthermore, both rule based and supervised systems typically operate in a pipeline approach where an NER engine identifies the relevant entities and subsequent extraction models identify the relations between them. This approach is often problematic in real world biomedical IE scenarios, where relevant entities often cannot be extracted by stock NER models.

\paragraph{To address these limitations} we present a system allowing domain experts to interactively query linguistically annotated datasets of scientific research papers, using a novel multifaceted query language which we designed, and which supports boolean search, sequential patterns search, and by-example syntactic search \cite{syntactic-search-2020}, as well as specification of search terms whose matches should be captured or expanded. The queries can be further restricted by contextual information.

We demonstrate the system on two datasets: a comprehensive dataset of PubMed abstracts and a dataset of full text papers focused on COVID-19 research.  \\
\noindent\textbf{Comparison to existing systems.} In contrast to document level search solutions, the results returned by our system are sentences which include highlighted spans that directly answer the user's information need. In contrast to supervised IE solutions, our solution does not require a lengthy process of data collection and labeling or a precise definition of the problem settings. 

Compared to rule based systems our system differentiates itself in a number of ways: (i) our query engine automatically translates lightly tagged natural language sentences to syntactic queries (query-by-example) thus allowing domain experts to benefit from the advantages of syntactic patterns without a deep understanding of syntax; (ii) our queries run against indexed data, allowing our translated syntactic queries to run at interactive speed; and (iii) our system does not rely on relation schemas and does not make assumptions about the number of arguments involved or their types. 

In many respects, our system is similar to the PropMiner system  \cite{propminer} for exploratory relation extraction \cite{akbik}. Both PropMiner and our system support by-example queries in interactive speed. However, the query languages we describe in section \ref{sec:languages} are significantly more expressive than PropMiner's language, which supports only binary relations. Furthermore, compared to PropMiner, our annotation pipeline was optimized specifically for the biomedical domain and our system is freely available online.\\
\noindent\textbf{Technical details.} The datasets were annotated for biomedical entities and syntax using a custom SciSpacy pipeline \cite{scispacy}\footnote{All abstracts underwent sentence splitting, tokenization, tagging, parsing and NER using all the 4 NER models available in SciSpacy}, and the syntactic trees were enriched to BART format using pyBART \cite{pyBART}. The annotated data is indexed  using the Odinson engine \cite{odinson}.

\section{Extractive Query Languages}
\label{sec:languages}
\subsection{Boolean Queries}
\label{sec:boolean}
Boolean queries are the standard in information retrieval (IR): the user provides a set of terms that should, and should not, appear in a document, and the system returns a set of documents that adhere to these constraints. This is a familiar and intuitive model, which can be very effective for initial data exploration as well as for extraction tasks that focus on co-occurrence. We depart from standard boolean queries and extend them by (a) allowing to condition on different linguistic aspects of each token; (b) allowing \emph{capturing} of terms into named variables; and (c) allowing \emph{linguistic expansion} of the captured terms.

The simplest boolean query is a list of terms, where each term is a word, i.e:
\query{\tgv{infection asymptomatic fatal}}
The semantics is that all the terms must appear in the query. A term can be made optional by prefixing it with a \query{\tgo{?}} symbol (\query{\tgv{infection asymptomatic} \tgo{?}\tgv{fatal}} ). Each term can also specify a list of alternatives: \query{\tgv{fatal}\tgo{$|$}\tgv{deadly}\tgo{$|$}\tgv{lethal}}.

\noindent\textbf{Beyond words.} In addition to matching words, terms can also specify linguistic properties: lemmas, parts-of-speech, and domain-specific entity-types: \query{\tgc{lemma}\tgo{=}\tgv{infect} \tgc{entity}\tgo{=}\tgv{DISEASE}}. Conditions can also be combined: \query{\tgc{lemma}\tgo{=}\tgv{cause}\tgo{$|$}\tgv{reason}\tgo{$\&$}\tgc{tag}\tgo{=}\tgv{NN}}. We find that the ability to search for domain-specific types is very effective in boolean queries, as it allows to search for concepts rather than words. In addition to exact match, we also support matching on regular expressions (\query{\tgc{lemma}\tgo{=}\tgv{/caus.*/}}). The field names \wquery{\tgc{word}, \tgc{lemma}, \tgc{entity}, \tgc{tag}} can be shortened to \wquery{\tgc{w},\tgc{l},\tgc{e},\tgc{t}}.

\noindent\textbf{Captures.} Central to our extractive approach is the ability to designate specific search term to be \emph{captured}. Capturing is indicated by prefixing the term with \query{\CAP} (for an automatically-named capture) or with \query{\tgn{name}\CAP} (for a named capture). The query\\
\query{\tgv{fatal asymptomatic} \tgn{d}\CAP{}\tgc{e}\tgo{=}\tgv{DISEASE}} will look for sentences that contain the terms `fatal` and `asymptomatic` as well as a name of a disease, and will capture the disease name under a variable ``d''. Each query result will be a sentence with a single disease captured. If several diseases appear in the same sentence, each one will be its own result. The user can then focus on the captured entities, and export the entire query result to a CSV file, in which each row contains the sentence, its source, and the captured variables. In the current examples, the result will be a list of disease names that co-occur with ``fatal'' and ``asymptomatic''. We can also issue a query such as\\ \query{\tgn{chem}\CAP{}\tgc{e}\tgo{=}\tgv{SIMPLE\_CHEMICAL}  \tgn{d}\CAP{}\tgc{e}\tgo{=}\tgv{DISEASE}} \\to get a list of chemical-disease co-occurrences. Using additional terms, we can narrow down to co-occurrences with specific words, and by using contextual restrictions (\S\ref{sec:restrictions}) we can focus on co-occurrences in specific papers or domains.

\noindent\textbf{Expansions.} Finally, for captured terms we also support \emph{linguistic expansions}. After the term is matched, we can expand it to a larger linguistic environment based on the underlying syntactic sentence representation. An expansion is expressed by prefixing a term with angle brackets \wquery{\tgo{$\expand$}}: 

\query{\tgo{$\expand$}\tgn{inf}\CAP{}\tgv{infection} \tgv{asymptomatic fatal}} 
\noindent will capture the word ``infection'' under the variable ``inf" and \emph{expand} it to its surrounding noun-phrase, capturing phrases like ``malarialike infection'',  ``asymptomatic infection'', ``chronic infection'' and ``a mild subclinical infection 9''.
\begin{figure*}[t]
\centering\includegraphics[width=0.8\textwidth]{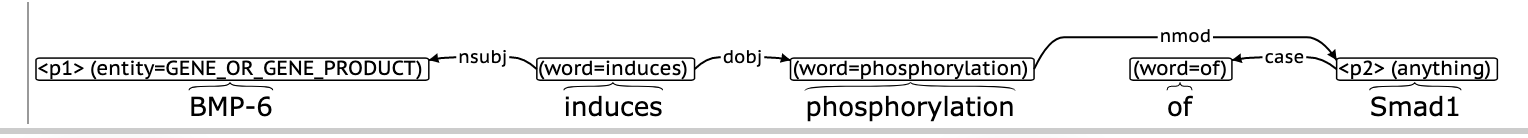}
\caption{Query Graph of the syntactic query \query{\tgo{$\expand$}\tgn{p1}\CAP\tgo{[}\tgc{e}\tgo{]}BMP-6 \tgo{\$}induces the \tgo{\$}phosphorylation \tgo{\$}of \tgo{$\expand$}\tgn{p2}\CAP{}Smad1.}}
\label{fig:graph_query}
\end{figure*}

\vspace{-0.5em}
\subsection{Sequential (Token) Queries}
\label{sec:sequential}
While boolean queries allow terms to appear in any order, we sometimes care about the exact linear placements of words with respect to each other. The term-specification, capture and expansion syntax is the same as in boolean queries, but here terms must match as in the query. \\
\query{\tgv{interspecies transmission}} looks for the exact phrase ``interspecies transmission'' and \\
\query{\tgc{tag}\tgo{=}\tgv{NNS} \tgv{transmission}} looks for the word transmission immediately preceded by a plural noun. By capturing the noun (\query{\tgn{which}\CAP{}\tgc{tag}\tgo{=}\tgv{NNS} \tgv{transmission}}) we obtain a list of terms that includes the words  ``bacteria'',  ``diseases'', ``nuclei'' and ``crossspecies''.

\noindent\textbf{Wildcards.} sequential queries can also use wildcard symbols: \tgv{*} (matching any single word), \query{\tgv{...}} (0 or more words), \query{\tgv{...2-5...}} (2-5 words). The query \query{\tgv{interspecies} \tgn{kind}\CAP{}\tgv{...1-3...} \tgv{transmission}} looks for the words ``interspecies'' and ``transmission'' with 1 to 3 intervening words, capturing the intervening words under ``kind''. First results  include  ``host-host'', ``zoonotic'', ``virus'', ``TSE agent'', ``and interclass''.

\noindent\textbf{Repetitions.} We also allow to specify repetitions of terms. To do so, the term is enclosed in \wquery{[ ]} and followed by a quantifier. We support the standard list of regular expression quantifiers: \wquery{\tgo{*}, \tgo{+}, \tgo{?}, \tgo{\{n,m\}}}. For example, \query{\tgc{tag}\tgo{=}\tgv{DT} \tgo{[}\tgc{tag}\tgo{=}\tgv{JJ}\tgo{]*} \tgo{[}\tgc{tag}\tgo{=}\tgv{NN}\tgo{]+}}.

\vspace{-0.5em}
\subsection{Contextual Restrictions}
\label{sec:restrictions}
Each query can be associated with contextual restrictions, which are secondary queries that operate on the same data and restrict the set of sentences that are considered for the main queries. These queries currently have the syntax of the Lucene query language.\footnote{\url{https://lucene.apache.org/core/6_0_0/queryparser/org/apache/lucene/queryparser/classic/package-summary.html}} Our system allows the secondary queries to condition on the paragraph the sentence appears in, and on the title, abstract, authors, publication data, publication venue and MeSH terms of the paper the sentence appears in. Additional sources of information are easy to add.  
For example, adding the contextual restriction \query{\tgo{\#d} \tgo{+}\tgc{title}\CAP{}\tgv{cancer} \tgo{+}\tgc{mesh}\CAP{}\tgo{"}\tgv{Age Distribution}\tgo{"}} restricts a query results to sentences from papers which have the word ``cancer" in their title and whose MeSH terms include ``Age Distribution". Similarly \query{\tgo{\#d} \tgo{+}\tgc{title}\CAP{}\tgv{/corona.*/} \tgo{+}\tgc{year}\CAP{} \tgo{[}\tgv{2015} \tgo{TO} \tgv{2020}\tgo{]}} restricts queries to include sentences from papers published between 2015 and 2020 and have a word starting with \emph{corona} in their title. 

These secondary queries greatly increase the power of boolean, sequential and syntactic queries: one could look for interspecies transmissions that relate to certain diseases, or for sentence-level disease-chemical co-occurrences in papers that discuss specific sub-populations.

\vspace{-5pt}
\subsection{Example-based Syntactic Queries}
\label{sec:query-by-example}
Recent advances in machine learning brought with them accurate syntactic parsing, but parse-trees remain hard to use. We remedy this by employing a novel query language we introduced in \cite{syntactic-search-2020} which is  based on the principle of query-by-example.

The query is specified by starting with a simple natural language sentence that conveys the desired syntactic structure, for example, `\emph{BMP-6 induces the phosphorylation of Smad1}'. Then, words can be marked as anchor words (that need to match exactly) or capture nodes (that are variables). Words can also be neither anchor or capture, in which case they only support the scaffolding of the sentence. The system then translates the sentence with the captures and anchors syntax into a syntactic query graph, which is presented to the user. The user can then restrict capture nodes from ``match anything" to matching specific terms (using the term specification syntax as in boolean or token queries) and can likewise relax the exact-match constraints on anchor words. Like in other query types, capture nodes can be marked for expansion.
The syntactic graph is then matched against the pre-parsed and indexed corpus.

This simple markup provides a rich syntax-based query system, while alleviating the user from the need to know linguistic syntax.

For example, consider the query below, the details of which will be discussed shortly:\\[3pt]
\query{\tgo{$\expand$}\tgn{p1}\tgo{:[}\tgc{e}\tgo{=}\tgv{GENE\_OR\_GENE\_PRODUCT}\tgo{]}BMP-6 \tgo{\$}induces the \tgo{\$}phosphorylation \tgo{\$}of \tgo{$\expand$}\tgn{p2}\CAP{}Smad1} 
\vspace{3pt}

The words `induce', `phosphorylation' and `of' are anchors (designated by \query{\tgo{\$}}), while `p1' and `p2' are captures for `BMP-6' and `Smad1'. Both capture nodes are marked for expansion using angle braces (\query{\tgo{$\expand$}}). Node p1 is \emph{restricted} to match tokens with the same entity type of BMP-6 (indicated by \query{\tgc{e}\tgo{=}\tgv{GENE\_OR...}}). 
The query can be shortened by omitting the entity type and retaining only the entity restriction (\query{\tgc{e}}):\\[3pt]
\query{\tgo{$\expand$}\tgn{p1}\CAP\tgo{[}\tgc{e}\tgo{]}BMP-6 \tgo{\$}induces the \tgo{\$}phosphorylation \tgo{\$}of \tgo{$\expand$}\tgn{p2}\CAP{}Smad1} 
\vspace{3pt}

Here, the entity type is inferred by the system from the entity type of BMP-6.\footnote{Similarly, we could specify \query{\tgo{\$[}\tgc{lemma}\tgo{]}induces}, resulting in the restriction `lemma=induce' instead of `word=induces' for the anchor.}
The graph for the query is displayed in Figure \ref{fig:graph_query}. It has 5 tokens in a specific syntactic configuration determined by directed labeled edges. The 1st token must have the entity tag of \query{\tgv{GENE\_OR...}}, the 2nd, 3rd, and 4th tokens must be the exact words ``induces phosphorylation of", and the 5th is unconstrained.

Sentences whose syntactic graph has a subgraph that aligns to the query and adheres to the constraints will match the query. Example of matching sentences are:\\
- \emph{\underline{ERK}$_{p_1}$ activation induces phosphorylation of  \underline{Elk-1}{$_{p_2}$}}. \\
- \emph{\underline{Thrombopoietin}$_{p_1}$ activates human platelets and induces tyrosine phosphorylation of \underline{p80/85 cortactin}$_{p2}$}\\[0.2em]
The sentence tokens corresponding to the p1 and p2 graph nodes will be bound to variables with these names: 
\wquery{\{p1=ERK, p2=Elk-1\}} for the first sentence and \wquery{\{p1=Thrombopoietin, p2=p80/85 cortactin\}} for the second.

\section{Example Workflow: Risk-factors}
\label{sec:example-workflow}
We describe a workflow which is based on using our extractive search system over a corpus of all PubMed abstracts. While the described researcher is hypothetical, the results we discuss are real.

    Consider a medical researcher who is trying to compile an up to date list of the risk factors for stroke. A PubMed search for ``risk factors for stroke" yields 3317 results, and reading through all results is impractical. A Google query for the same phrase brings out an info box from NHLBI\footnote{\url{https://www.nhlbi.nih.gov/health-topics/stroke}} listing 16 common risk factors including high blood pressure, diabetes, heart disease, etc. 
    Having a curated list which clearly outlines the risk factors is helpful, but curated lists or survey papers will often not include rare or recent research findings. 
    
    The researcher thus turns to extractive search and tries an exploratory boolean query:
    \\[0.5em]
    \query{\tgv{risk factor stroke}}
    \noindent\includegraphics[width=0.5\textwidth]{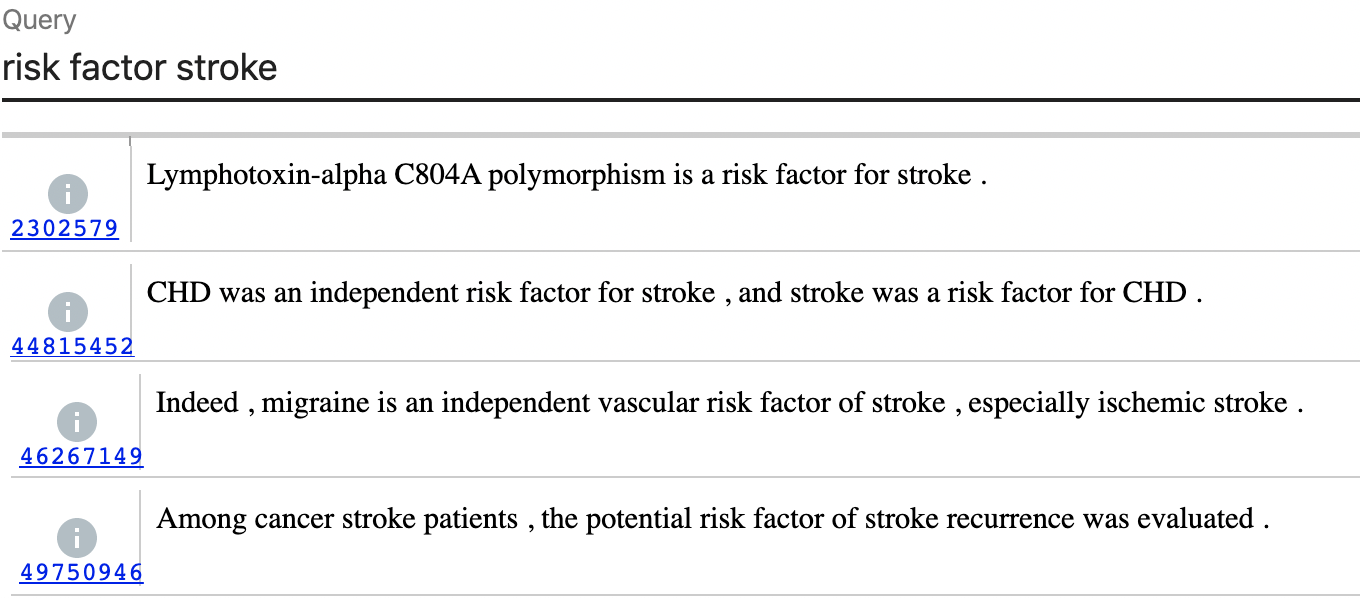}. 
    
    The figure shows the top results for the query and the majority of sentences retrieved indeed specify specific risk factors for stroke. This is an improvement over the PubMed results as the researcher can quickly identify the risk factors discussed without going through the different papers. 
    
    Furthermore, the top results contain risk factors like \emph{migrane} or \emph{C804A polymorphism} not listed in the NHLBI knowledge base. However, the full result list is lengthy and extracting all the risk factors from it manually would be tedious. Instead, the researcher notes that many of the top results are variations on the ``X is a risk factor for stroke" structure. She thus continues by issuing the following syntactic query, where a capture labeled \emph{r} is used to directly capture the risk factors:
    \\[0.5em]
    \query{\tgn{r}\CAP{}Diabetes is a \tgo{\$}risk \tgo{\$}factor for \tgo{\$}stroke}. 
    
    \noindent\includegraphics[width=0.5\textwidth]{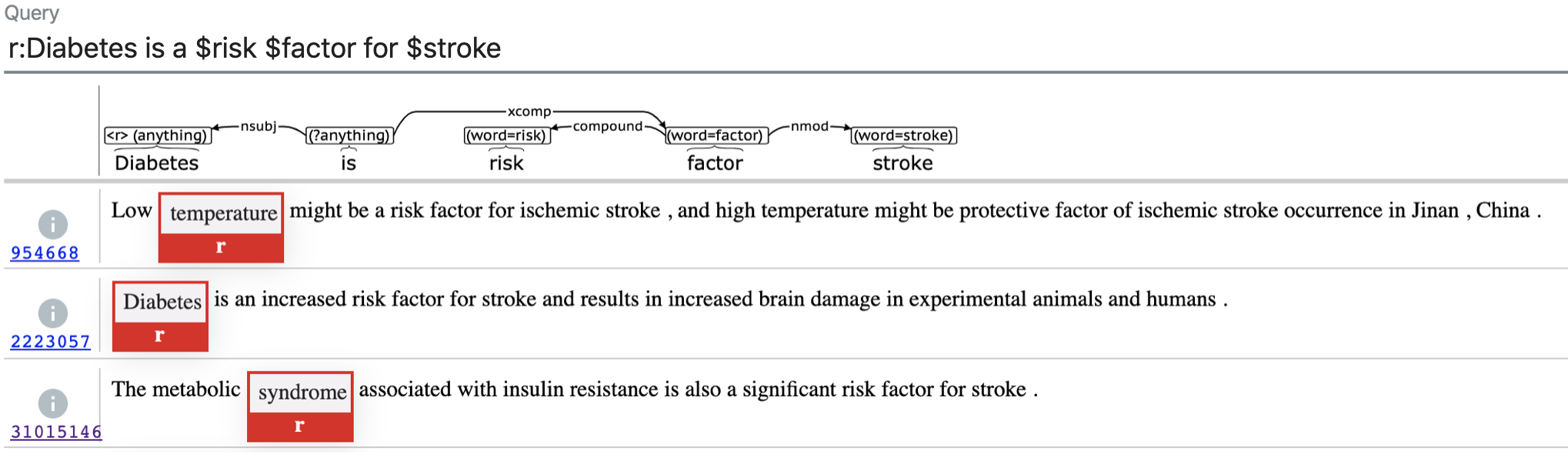}
    
    The figure 
    shows the top results for the query and the risk factors are indeed labeled with \emph{r} as expected. 
    Unfortunately, some of the captured risk factors names are not fully expanded. For example, we capture \emph{syndrome} instead of \emph{metabolic syndrome} and \emph{temperature} instead of \emph{low temperature}. Being interested in capturing the full names, the researcher adds angle brackets \query{\tgo{$\expand$}} to expand the captured elements:
    \\[0.5em]
    \query{\tgo{$\expand$}\tgn{r}\CAP{}Diabetes is a \tgo{\$}risk \tgo{\$}factor for \tgo{\$}stroke}. 
    
    \noindent\includegraphics[width=0.5\textwidth]{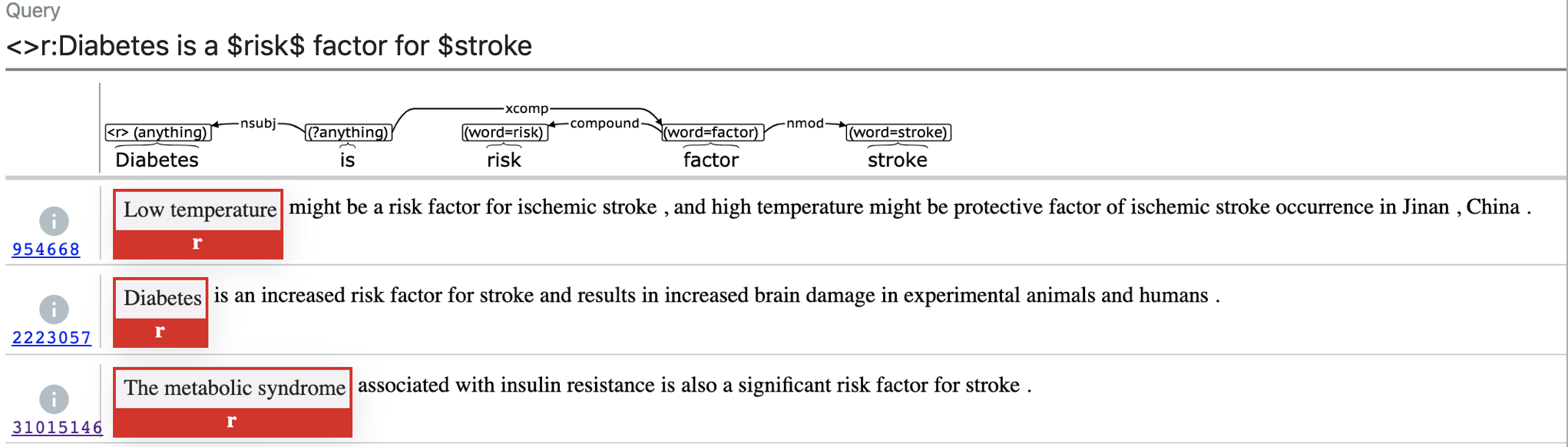}
    
    \noindent The full names are now captured as expected. 
    
    Now that that researcher has verified that the query yields relevant results, she clicks the download button to download the full result set.
    
\begin{figure}
    \begin{subfigure}{.25\textwidth}
        \includegraphics[width=.7\linewidth]{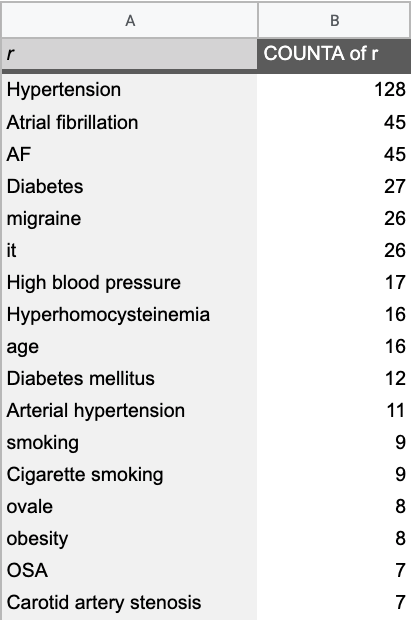}
        \caption{ranked risk factors for stroke}
        \label{fig:risk_factors_ranked}
    \end{subfigure}%
    \begin{subfigure}{.25\textwidth}
        \includegraphics[width=.7\linewidth]{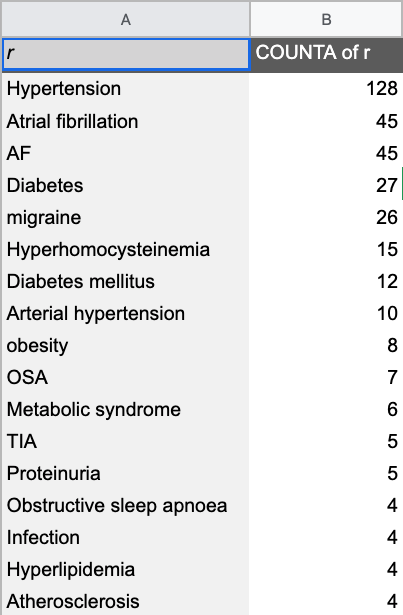}
        \caption{ranked disease risk factors}
        \label{fig:disease_factors_ranked}
    \end{subfigure}
    \caption{Grouped and ranked results}
    \end{figure}

    The resulting tab separated file 
    has 1212 rows. Each row includes a result sentence, the captured elements in it (in this case, just the risk factor), and their offsets. Using a spreadsheet to group the rows by risk factor and order the results by frequency, the researcher obtains a list of 640 unique risk factors, 114 of them appearing more than once in the data. Figure \ref{fig:risk_factors_ranked} lists the top results.

    Reviewing the list, the researcher decides that she's not interested in general risk factors, but rather in diseases only. She modifies the query by adding an entity restriction to the `r' capture:

    \noindent\includegraphics[width=0.5\textwidth]{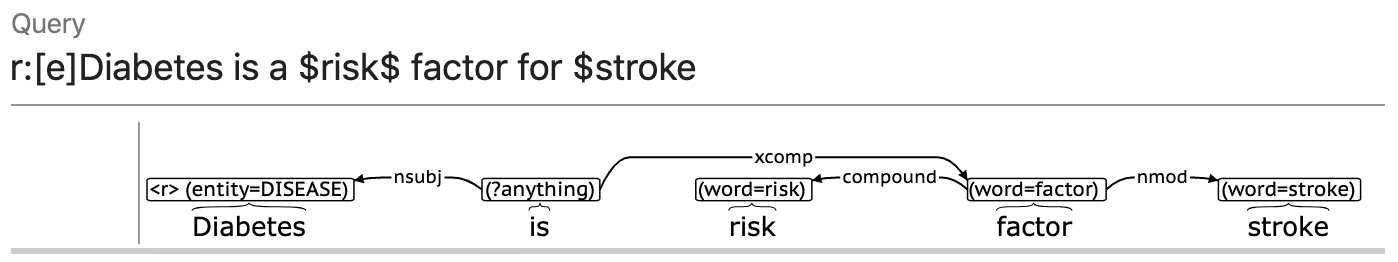}

    As seen in 
    the query graph, even though the researcher didn't specify the exact entity type, the query parser correctly resolved it to DISEASE.
    The results now include diseases like sleep apnoea and hypertension but do not include smoking, age and alcohol (see Figure \ref{fig:disease_factors_ranked}). 
    
    Analyzing the results, the researcher now wants to compare the risk factors in the general population to ones listed in research papers dealing with children and infants. Luckily, such papers are indexed with corresponding MeSH terms and the researcher can utilize this fact by appending \query{\tgo{\#d} \tgc{mesh}\CAP{}\tgv{Child} \tgc{mesh}\CAP{}\tgv{Infant} \tgo{-}\tgc{mesh}\CAP{}\tgv{Adult}} to her query. In cases where a desired MeSH term does not exist, an alternative approach is filtering the results based on words in the abstract or title. For example, appending \query{\tgo{\#d} \tgc{abstract}\CAP{}\tgv{child} \tgc{abstract}\CAP{}\tgv{children}} to a query will ensure that the result sentences come from abstracts which contain the word \emph{child} or the word \emph{children}.
    
    Happy with the results of the initial query, the researcher can further augment her list by querying for other structures which identify risk factors (e.g. ``\query{\tgn{r}\CAP{}Diabetes \tgo{\$}causes \tgo{\$}stroke}'', ``\query{\tgo{\$}risk \tgo{\$}factors for \tgo{\$}stroke \tgo{\$}include \tgn{r}\CAP{}Diabetes}'', etc.). 
    
    Importantly, once the researcher has identified one or more effective queries to extract the risk factors for stroke, the queries can easily be modified in useful ways. For example, with a small modification to our original query we can extract:\\
    \textbf{risk factors for cancer:}\\
   \indent\query{\tgn{r}\CAP{}Diabetes is a \tgo{\$}risk \tgo{\$}factor for \tgo{\$}cacner}  \\[0.2em]
    \textbf{diseases which can be caused by smoking:}\\
    \indent\query{\tgo{\$}Smoking is a \tgo{\$}risk \tgo{\$}factor for \tgn{d}\tgo{:[}\tgc{e}\tgo{]}stroke}. \\[0.2em]
    \textbf{ad-hoc KB of (risk factor, disease) tuples} (for self use or as an easily queryable public resource):\\
    \indent\query{\tgn{r}\CAP{}Diabetes is a \tgo{\$}risk \tgo{\$} factor for \tgn{d}\tgo{:[}\tgc{e}\tgo{]}stroke}.
   
\vspace{-0.5em}
\section{Example Workflow: CORD-19}
The COVID-19 Open Research Dataset \cite{cord-19} is a collection of 45,000 research papers, including over 33,000 with full text, about COVID-19 and the coronavirus family. The corpus was released by the Allen Institute for AI and associated partners in an attempt to encourage researchers to apply recent advances in NLP to the data to generate insights.

\paragraph{Identifying COVID-19 Aliases}
Since the CORD-19 corpus includes papers about the entire Coronavirus family of viruses, it's useful to identify papers and sentences dealing specifically with COVID-19. Before converging on the acronym COVID-19 researchers have referred to the virus by many names: nCov-19, SARS-COV-ii, novel coronavirus, etc. Luckily, it's fairly easy to identify many of these aliases using a sequential pattern:

\query{\tgv{novel coronavirus} \tgo{(} \tgn{alias}\CAP{}\tgv{...1-2...} \tgo{)}}

\noindent\includegraphics[width=0.5\textwidth]{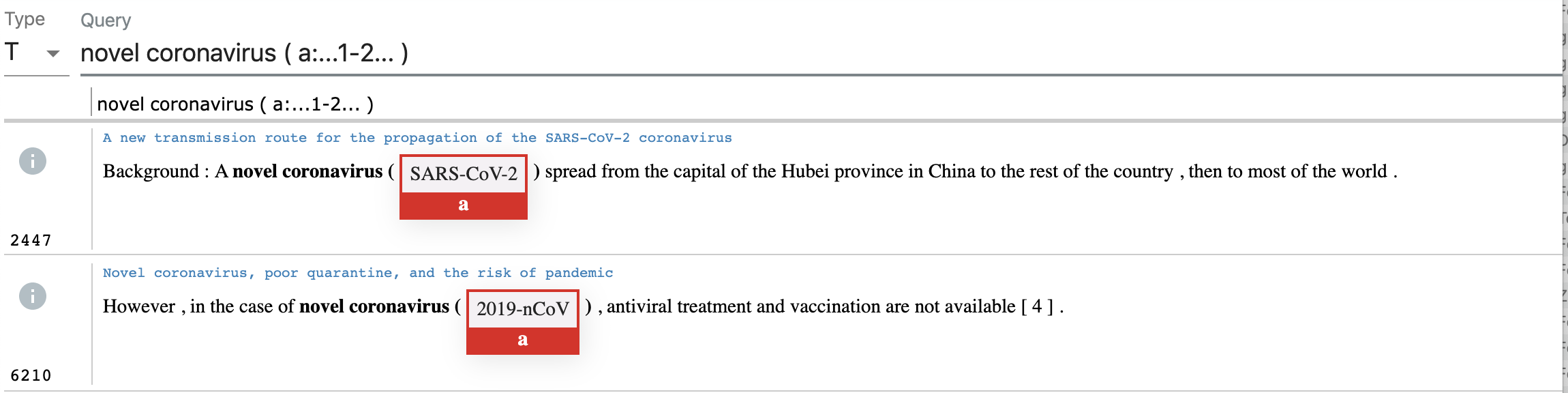}

\noindent The pattern looks for the words ``novel coronavirus'' followed by an open parenthesis, one-or-two words which are to be captured under the `alias' variable, and a closing parenthesis. The query retrieves 52 unique candidate aliases for COVID-19, though some of them refer to older coronaviruses such as ``MERS'', or non-relevant terms such as ``Fig2''. 
After ranking by frequency and validating the results, we can reuse the pattern on newly retrieved aliases to extend the list. Through this iterative process we quickly compile a list of 47 aliases. We marked all occurrences of these terms in the underlying corpus as a new entity type, COVID-19, and re-indexed the dataset with this entity information. 

\begin{table}[t]
\hrule
\vspace{3pt}
(a) Unrestricted
\vspace{3pt}
\hrule
\vspace{3pt}
{\footnotesize nucleic acid (171), chloroquine (118), nucleotide (115), NCP (87), CR3022 (47), Ksiazek (46), IgG (45), lopinavir/ritonavir (42), ECMO (40), LPV/r (35), corticosteroids (35), oxygen (32), ribavirin (31), lopinavir (31), Hydroxychloroquine (30), amino acid (30), ritonavir (27), corticosteroid (24), Sofosbuvir (22), amino acids (22), HCQ (19), glucocorticoids (19)}
\vspace{3pt}
\hrule width \hsize \kern 0.5mm \hrule width \hsize 
\vspace{3pt}
(b) Trial
\vspace{3pt}
\hrule
\vspace{3pt}
{\footnotesize chloroquine (29), Remdesivir (8), LPV/r (7), lopinavir (6), HCQ (6), ritonavir (4), Arbidol (4), Sofosbuvir (3), nucleotide (3), nucleic acid (3), lopinavir/ritonavir (3), CQ (3), oseltamivir (2), NCT04257656 (2), NCT04252664 (2), Meplazumab (2), Hydroxychloroquine (2), glucocorticoids(2), CEP(2)}
\vspace{3pt}
\hrule width \hsize \kern 0.5mm \hrule width \hsize 
\vspace{3pt}
(c) Ideation
\vspace{3pt}
\hrule
\vspace{3pt}
{\footnotesize chloroquine (6), ritonavir (5),	S-RBD (4), nucleotide (4), Lopinavir (4), CR3022 (4), Ribavirin (3), nucleic acid (3), logP (3), Li (3), ledipasvir (3), IgG (3), HCQ (3), TGEV (2), teicoplanin (2), nelfinavir (2), NCP (2), HWs (2)	glucocorticoids (2), ENPEP (2), ECMO (2), darunavir (2), creatinine (2), creatine (2), CQ (2), corticosteroid (2), CEP (2), ARB (2)}
\vspace{3pt}
\hrule width \hsize \kern 0.5mm \hrule width \hsize 
\vspace{3pt}
\caption{Top chemicals co-occuring with the COVID-19 entity and their counts. (a) Unrestricted. (b) with Trial related terms. (c) with Ideation related terms.}
\label{tbl:treatments}
\vspace{-1em}
\end{table}

\noindent\textbf{Exploring Drugs and Treatments.}
To explore drugs and treatments for COVID-19 we search the corpus for chemicals co-occuring with the COVID-19 entity using a boolean query:\\
\query{\tgn{chemical}\CAP{}\tgc{e}\tgo{=}\tgv{SIMPLE\_CHEMICAL}\tgo{$|$}\tgv{CHEMICAL} \tgc{e}\tgo{=}\tgv{COVID-19}}

Table \ref{tbl:treatments}(a) shows the top matching chemicals by frequency. While some of the substances listed like \emph{Chloroquine} and \emph{Remdesivir} are drugs being tested for treating COVID-19, others are only hypothesized as useful or appear in other contexts. 

To guide the search toward therapeutic substances in different stages of maturity we can add indicative terms to the query. For example, the following query can be used to detect substances at the stage of clinical trials:\\
\query{\tgn{chemical}\CAP{}\tgc{e}\tgo{=}\tgv{SIMPLE\_CHEMICAL}\tgo{$|$}\tgv{CHEMICAL} \tgc{e}\tgo{=}\tgv{COVID-19} \tgc{l}\tgo{=}\tgv{trial}\tgo{$|$}\tgv{experiment}},
while adding \query{\tgc{l}\tgo{=}\tgv{suggest}\tgo{$|$}\tgv{hypothesize}\tgo{$|$}\tgv{candidate}}
can assist in detecting substances in ideation stage.

Table \ref{tbl:treatments}(b,c) shows the frequency distributions of the chemicals resulting from the two queries. While the queries are very basic and include only a few terms for each category, the difference is clearly noticeable: while the Malaria drug Chloroquine tops both lists, the antiviral drug Remdesivir which is currently tested for COVID-19 is second on the list of trial related drugs but does not appear at all as a top result for ideation related drugs.

Importantly, entity co-mention queries like the ones above rely on the availability and accuracy of underlying NER models. As we've seen in Section \ref{sec:example-workflow}, in cases where the relevant types are not extracted by NER, syntactic queries can be used instead. For example the following query captures sentences including chemicals being used on patients (the abstract or paragraph are required to include COVID-19 related terms). \\
\query{he was \tgo{\$}treated \tgo{\$}with a \tgo{$\expand$}\tgn{chem}\CAP{}treatment \\\tgo{\#d} \tgc{paragraph}\CAP{}\tgv{ncov*} \tgc{paragraph}\CAP{}\tgv{covid*} \tgc{abstract}\CAP{}\tgv{ncov*} \tgc{abstract}\CAP{}\tgv{covid*}}\\[0.2em]
\noindent\includegraphics[width=0.5\textwidth]{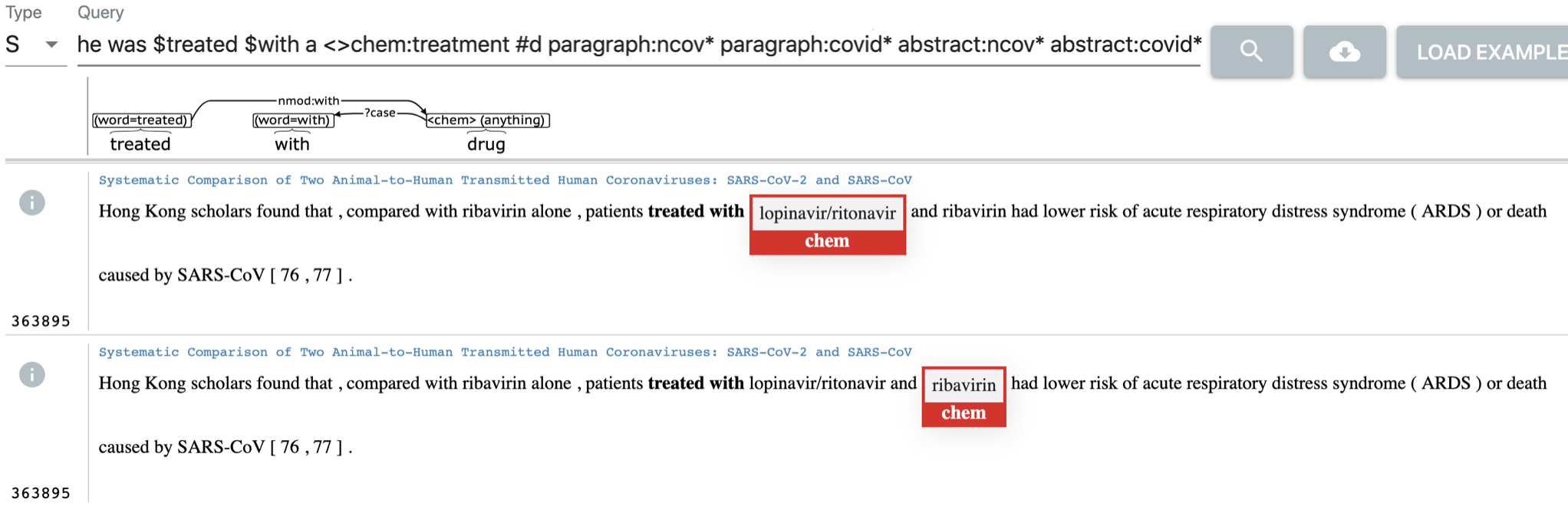}

The top results by frequency are shown in Table \ref{tbl:syn_treatments}. The top ranking results show many of the chemicals obtained by equivalent boolean queries\footnote{to get a more comprehensive coverage we can issue queries for other syntactic structures like \query{\tgo{$\expand$}\tgn{chem}\CAP{}chemical was used \tgo{\$}in \tgo{\$}treatment} and combine the results of the different queries.}, but interestingly, they also contain non-chemical treatments like \emph{supportive care}, \emph{isolation} and \emph{masks}. This demonstrates a benefit of using entity agnostic syntactic patterns even in cases where a strong NER model exists.

\begin{table}[t]
\hrule
\vspace{3pt}
Treatments (via syntactic query)
\vspace{3pt}
\hrule
\vspace{3pt}
{\footnotesize ribavirin (11), oseltamivir (9), ECMO (6), convalescent plasma (4), TCM (3), LPV/r (3), three fusions of MSCs (2), supportive care (2), protective conditions (2), lopinavir/ritonavir (2), intravenous remdesivir (2), hydroxychloroquine (2), HCQ (2), glucocorticoids (2), FPV (2), effective isolation (2), chloroquine (2), caution (2), bDMARDs (2), azithromycin (2), ARBs (2), antivirals (2), ACE inhibitors (2), 500 mg chloroquine (2), masks (1)}
\vspace{3pt}
\hrule width \hsize \kern 0.5mm \hrule width \hsize 
\vspace{3pt}
\caption{Top elements occurring in the syntactic ``treated with X" configuration. Note that this query does not rely on NER information.}
\label{tbl:syn_treatments}
\vspace{-1em}
\end{table}

\section{More Examples}
While the workflows discussed above pertain mainly to the medical domain, the system is optimized for the broader life science domain. Here are a sample of additional queries, showing different potential use-cases.

\noindent\textbf{Which genes regulate a cell process}:\\
\noindent\query{\tgo{$\expand$}\tgn{p1}\tgo{:[}\tgc{e}\tgo{]}CD95 \tgn{v}\tgo{:[}\tgc{l}\tgo{]}regulates \tgo{$\expand$}\tgn{p}\tgo{:[}\tgc{e}\tgo{]}apoptosis}

\noindent\textbf{Which specie is the natural host of a disease}:\\
\noindent\query{\tgo{$\expand$}\tgn{host}\tgo{:[}\tgc{e}\tgo{]}bat is a \tgo{\$}natural \tgo{\$}host of \tgo{$\expand$}\tgn{disease}\tgo{:[}\tgc{e}\tgo{]}coronavirus}

\noindent\textbf{Documented LOF mutations in genes}:\\
\noindent\query{\tgo{\$}loss \tgo{\$}of \tgo{\$}function \tgo{$\expand$}\tgn{m}\tgo{:[}\tgc{w}\tgo{]}mutation in \tgo{$\expand$}\tgn{gene}\tgo{:[}\tgc{e}\tgo{]}PAX8}

\section{Conclusion}
We presented a search system that targets extracting facts from a biomed corpus and demonstrated its utility in a research and a clinical context over CORD-19 and PubMed. The system works in an Extractive Search paradigm which allows rapid information seeking practices in 3 modes: boolean, sequential and syntactic. The interactive and flexible nature of the system makes it suitable for users in different levels of sophistication.

\paragraph{Acknowledgements}
The work performed at BIU is supported by funding from the Europoean Research Council (ERC) under the Europoean Union's Horizon 2020 research and innovation programme, grant agreement No. 802774 (iEXTRACT).

\bibliography{anthology,acl2020,local}
\bibliographystyle{acl_natbib}

\end{document}